\documentclass[sigconf]{acmart}
\let\svthefootnote\thefootnote

\usepackage[utf8]{inputenc}
\usepackage{graphicx}
\usepackage{tabularx}
\usepackage{multirow}
\usepackage{todonotes}
\usepackage{amsmath}
\usepackage[misc,geometry]{ifsym}
\usepackage{xcolor}
\usepackage{todonotes}
\usepackage{array} % for defining a new column type
\newcolumntype{P}[1]{>{\centering\arraybackslash}p{#1}}
\newcolumntype{M}[1]{>{\centering\arraybackslash}m{#1}}

\NewDocumentCommand{\rot}{O{45} O{1em} m}{\makebox[#2][l]{\rotatebox{#1}{#3}}}%
\usepackage{ltablex}
\usepackage{graphicx}
\usepackage{cleveref}
\usepackage{enumitem,kantlipsum}
\usepackage{multirow}
\usepackage{siunitx}

\usepackage{color,soul}

\usepackage{caption}
\usepackage{subcaption}

\def\BibTeX{{\rm B\kern-.05em{\sc i\kern-.025em b}\kern-.08emT\kern-.1667em\lower.7ex\hbox{E}\kern-.125emX}}
\copyrightyear{2021}
\acmYear{2021}
\setcopyright{acmlicensed}
\begin{document}

% \title[LLMs for Complex Relevance Assessment]{Is ChatGPT a Good Assessor for Complex Relevance Tasks?\\Exploring the use of Large Language Models while Building Collections of Sentences Related to Depression Symptoms}

\title[DepreSym: A Depression Symptom Annotated Corpus]{DepreSym: A Depression Symptom Annotated Corpus and the Role of LLMs as Assessors of Psychological Markers}

%\title{\textbf{Is ChatGPT a Good Assessor for Complex Relevance?\\The Case of Pooled Datasets for Depression Symptoms Sentences}}

%%% Sugerencia de título: DepreSym: the Construction of a Pooled Dataset for Depression Symptoms Sentences using/testing ChatGPT as Assessor

%
% The "author" command and its associated commands are used to define the authors and their affiliations.
% Of note is the shared affiliation of the first two authors, and the "authornote" and "authornotemark" commands
% used to denote shared contribution to the research.
% \author{Marcos Fernández-Pichel, David E. Losada and Juan C. Pichel}
% \affiliation{%
%   \institution{Centro Singular de Investigación en Tecnoloxías Intelixentes (CiTIUS), Universidade de Santiago de Compostela}
%   \streetaddress{Rúa Jenaro de la Fuente s/n}
%   \city{15782 Santiago de Compostela}
%   \state{Spain}
% }
% \email{{marcosfernandez.pichel, david.losada, juancarlos.pichel}@usc.es}
%
% By default, the full list of authors will be used in the page headers. Often, this list is too long, and will overlap
% other information printed in the page headers. This command allows the author to define a more concise list
% of authors' names for this purpose.

\author{Anxo Pérez, Javier Parapar}
\email{{anxo.pvila, javier.parapar}@udc.es}
\affiliation{%
  \institution{Information Retrieval Lab, CITIC \\
  Universidade da Coruña}
  % \streetaddress{P.O. Box 1212}
  % \city{Dublin}
  \city{A Coruña}
  \country{Spain} 
  % \state{Ohio}
  % \country{USA}
  % \postcode{43017-6221}
}

\author{Marcos Fernández-Pichel, David E. Losada}
\email{{marcosfernandez.pichel, david.losada}@usc.es}

\affiliation{%
  \institution{Centro Singular de Investigaci\'on en Tecnolox\'{\i}as Intelixentes (CiTIUS) \\
  Universidade de Santiago de Compostela}
  % \streetaddress{P.O. Box 1212}
  % \city{Dublin}
  % \postcode{15782}
  \city{Santiago de Compostela, 15782}
  \country{Galicia, Spain} 
  % \state{Galicia}
  % \country{USA}
}

\renewcommand{\shortauthors}{Anxo Pérez, Marcos Fernández-Pichel, Javier Parapar and David E. Losada}

\begin{abstract}

% Our paper represents a step forward...
% The principal beneficiaries of this study are healthcare professionals, helping in a global health task that cannot be fully covered by clinicians...
% Make emphasis in that we compared GPT-4 aswell

% Several works explored proxy-based methods for automatically annotating the mental health status of individuals in social media. Based on these binary annotations (depressive vs. control users), mental health prediction models aim to label users with published traces of depression based on their online information. 

Computational methods for depression detection aim to mine traces of depression
from online publications posted by Internet users. However, solutions trained on existing collections exhibit limited generalisation and interpretability. To tackle these issues, recent studies have shown that identifying depressive symptoms can lead to more robust models. The eRisk initiative fosters research on this area and
has recently proposed a new ranking task focused on developing search methods to find sentences related to depressive symptoms. This search challenge relies on the symptoms specified by the Beck Depression Inventory-II (BDI-II), a questionnaire widely used in clinical practice. Based on the participant systems' results, we present the \textit{DepreSym} dataset, consisting of \num{21580} sentences 
annotated according to their relevance to the 21 BDI-II symptoms. The labelled sentences come from a pool of diverse ranking methods, and the final dataset serves as a valuable resource for advancing the development of models that incorporate depressive markers such as clinical symptoms. Due to the complex nature of this relevance annotation, we designed a robust assessment methodology carried out by three expert assessors (including an expert psychologist). Additionally, we explore here the feasibility of employing recent Large Language Models (ChatGPT and GPT4) as potential assessors in this complex task. We undertake a comprehensive examination
of their performance, determine their main limitations and analyze
their role as a complement or replacement for human annotators. 

% We compare their performance, limitations, and differences with human annotations, and discuss whether they can complement traditional human annotation methods.

% The result is this task is the \textit{DepreSym} resource, 

% However, recent literature demonstrated their limitations in terms of generalization and interpretation.  

\end{abstract}


\settopmatter{printacmref=false}

\maketitle
\let\thefootnote\relax\footnotetext{"Copyright © 2020 for this paper by its authors. Use permitted under Creative Commons License Attribution 4.0 International (CC BY 4.0)."}
\addtocounter{footnote}{0}\let\thefootnote\svthefootnote

\section{Introduction}

Inspired by clinical practice, there has been a growing interest in designing predictive models that focus on identifying depressive symptoms~\cite{DBLP:conf/ijcai/ZhangCWZ22,PEREZ2022102380,nguyen-etal-2022-improving,Ríssola2022}. These approaches diverge from traditional depression screening models that rely on the presence of general markers, which are based on the use of engineered features (e.g., word counts, emotion levels, posting hours). However, these features offer less personalized and interpretable solutions~\cite{harrigian2020models, walsh2020stigma}. Based on this idea, recent studies have shown the potential of symptom-based detection models ~\cite {PEREZ2022102380,DBLP:conf/ijcai/ZhangCWZ22,perez2022semantic,nguyen-etal-2022-improving}. 

In this context, the eRisk depression severity task~\cite{losada2019overview} released the first collection that contains symptom-level input obtained directly from individuals. The goal of the task was to estimate the severity of the 21 symptoms 
enumerated in the BDI-II~\cite{beck1996comparison}. The BDI-II is a standardized questionnaire containing symptoms such as pessimism, suicidal ideas or sleep problems. Each symptom is handled by a question, which has four possible closed-text answers. The answers are presented in an ordinal scale representing the symptom severity. The eRisk's organizers designed a \textit{human-in-the-loop} approach, 
where real responses to the BDI-II were provided by Reddit users. These users also gave consent to access their publications' history on the platform. This innovative challenge aimed to foster the design of automatic
systems capable to analyze the thread of users' messages
and predict their responses to the BDI-II. 
However, the ground truth (users' responses to the BDI-II) was at user level. Thus, there was no explicit association between BDI-II symptoms and specific textual extracts from the publications. 

Two recent studies have attempted to fill this void by constructing fine-grained datasets that label depressive symptoms at sentence level~\cite{zhang-etal-2022-symptom,perez2023sigir}. This paper contributes to this line of research by introducing \textit{DepreSym}, a dataset to encourage the development of models that rely on symptom-level screening of depression. \textit{DepreSym} consists of \num{21580} sentences that are 
labeled in terms of their relevance to the BDI-II symptoms\footnote{\url{https://erisk.irlab.org/depresym\_dataset.html}}. This resource comes from a shared-data ranking task introduced in the CLEF 2023 eRisk Lab\footnote{\url{https://early.irlab.org/}}. To construct our dataset, three expert assessors annotated a pool of sentences associated with each symptom.
The candidate sentences were obtained using
top-k pooling from the relevance rankings 
designed by the participants in the task, with a total of $37$ different ranking methods presented. Pooling over the results of the participants helps in increasing the diversity of the candidate sentences.

The assessors were instructed to consider the candidate sentences as relevant if they were on-topic but also provide explicit information about the individual state related to
the symptom. This two-side notion of relevance is more complex compared to previous works, requiring us to develop a robust annotation methodology with formal assessment guidelines. To validate the effectiveness of our methodology, we calculate the inter-rater agreement and conduct further analysis of the resulting set of judgements. Additionally, we explore the ability of recent state-of-the-art Large Language Models (LLMs) to annotate the dataset. Specifically, we employ the latest versions of GPT conversational applications (ChatGPT \cite{chatgpt} and GPT-4 \cite{gpt4}) as complex relevance assessors. One of the main advantages of LLMs
is their ability to accurately process large amounts of data, which can significantly reduce the time and effort required for manual assessment. Comparing the performance of LLMs with human assessors provides insights into the strengths and limitations of both approaches. Human assessors are considered the gold standard for relevance assessment, but they are also subject to biases and errors that can affect their performance. Examining the performance of LLMs in relation to humans can help us to understand how well these models can replicate human behaviour. 
 
% During the annotation process, we included a brief meeting where the annotators discussed some sentences that were in doubt. This approach aimed to enhance the annotation quality and improve the inter-rater agreement. To evaluate the effectiveness of this approach, we compared the quality of the annotations with and without the briefing session. The results showed that the annotation quality and inter-rater agreement improved significantly after the annotators discussed the uncertain sentences. This approach helped to minimize subjectivity and enhance the reliability of the annotation process.

%\textbf{Aqui remarcar diferencias con datasets anteriores}

% Recently, large language models (LLMs) such as GPT-3 have shown remarkable performance in various NLP tasks \cite{brown2020language}. However, whether these models can be used as good assessors of complex relevance concepts remains unknown.

\section{Related work}
%%% Comparar con datasets anteriores? -> RW

\subsection{Depression Symptoms and Social Media}

Recent studies have incorporated the detection of fine-grained depressive symptoms to improve mental health models
and demonstrated their potential to improve  performance, generalizability and interpretability~\cite{PEREZ2022102380,DBLP:conf/ijcai/ZhangCWZ22,perez2022semantic,nguyen-etal-2022-improving}. Nevertheless, due to the lack of datasets containing symptom-level annotations, these methods were compelled to employ unsupervised learning solutions. Some researchers have followed clinical standards, such as those established in the BDI-II or the 9-Question Patient Health Questionnaire (PHQ-9)~\cite{kroenke2001phq}, to design robust heuristics to train symptom detection classifiers~\cite{coppersmith2018natural,DBLP:conf/ijcai/ZhangCWZ22,DBLP:conf/acl/NguyenYZDC22,PEREZ2022102380}. Mowery et al. also proposed an annotation scheme for Twitter data based on DSM-5 depression criteria~\cite{mowery2015towards}.

Two recent studies released resources that contain 
markers of depression symptoms at the sentence level. The PsySym dataset~\cite{zhang-etal-2022-symptom} is the first annotated sentence dataset covering 38 symptoms related to 7 mental disorders. The authors followed the DSM-5 guidelines to retrieve and annotate candidate sentences. In a recent contribution by Perez et al.~\cite{perez2023sigir}, the eRisk2019 collection was leveraged to obtain representative sentences for each BDI-II symptom. As a result, the authors released a dual dataset consisting of representative sentences for each symptom and the users' responses to the BDI-II.
We contribute here to this body of research by
releasing a large dataset of sentences that are related to the 21 BDI-II symptoms. Additionally, in the current work, the annotation process followed a robust methodological approach that involved a variety of retrieval systems to nominate candidate sentences presented to human assessors, including an expert psychologist.

\subsection{LLMs as Relevance Annotators} %%% TO-DO: Refactorizar 

High-quality assessments are essential to obtain accurate and reliable results~\cite{buttcher2007reliable},
and annotations need to be consistent, unbiased, and representative of the task at hand. Low-quality assessments potentially lead to inaccurate evaluations and unreliable conclusions~\cite{scholer2011quantifying}. The process of manually annotating test collections requires significant human effort, frequently requiring domain experts. As a consequence, a number of steps have been taken to reduce the cost and biases of the labelling process~\cite{moghadasi2013low,sakai2009robustness}. 

With the incredible development of LLMs, a potential application of these models is to assist in tasks such as relevance labelling. This represents a natural advance, as was the replacement of TREC annotators by crowdsourcing~\cite{alonso2009can}. 
% me está molando este refactor, si señor estas inspirado. Good job
% hahahah 
Initial steps were taken by Gilardi and his colleagues~\cite{gilardi2023chatgpt} demonstrating that ChatGPT outperforms crowd-workers for a tweet annotation task. Other researchers focused their efforts on improving their performance as annotators through prompt engineering~\cite{he2023annollm}. Faggiogli et al.~\cite{faggioli2023perspectives} tested the accuracy of LLMs for annotating two TREC test collections. Previous efforts by Meyer et al. \cite{meyer2022we} 
were oriented to produce synthetic training data for a conversational agent in the context of behaviour change. In this study, we intend to go one step further by evaluating the most recent LLMs for a highly demanding annotation task. Specifically, we put them under scrutiny for assessing the relevance
of sentences given specific BDI-II symptoms. Thus, we are considering a scenario where the two-side relevance notion is more complex (i.e., on-topic and providing explicit information about the individual). Moreover, the context is much shorter (i.e., only judging short sentences). To study this effect, we analyse the agreement between human annotations, including those coming from experts in the field, and machine annotations.

\section{Resource} 
\label{sec:resource} 

\begin{table}[t]
        \footnotesize
        \centering
        \caption{Examples of sentences for the symptom \textit{Loss of Energy}. Sentences are paraphrased for anonymity purposes. }
        \begin{tabular}{cl}
        \textbf{Relevance} & \multicolumn{1}{c}{\textbf{Sentence}} \\
        \midrule 
        \textbf{\multirow{2}*{0}} & ``\textit{Learn new ideas consumes energy, but builds neural connections.}''\\
        & 	``\textit{Low electrolytes can cause a person to feel low on energy.}''\\
        \midrule
        \textbf{\multirow{2}*{1}} & ``\textit{Even brushing my teeth is too exhausting for me right now.}''\\        
         & ``\textit{I became constantly lethargic, drowsy, and unable to concentrate.}''\\ \hline
        \end{tabular}
        \label{tab:examples_relevant_sentences}
\end{table}

% % [11:24, 26/4/2023] David Losada USC: es decir que son X sujetos de previas colecciones de erisk, que se segmentaron sus textos a nivel de oración y se generó un corpus de Y oraciones.
% [11:25, 26/4/2023] David Losada USC: lo que pasa que a ver cómo contamos, javier, el tema de pooling sin revelar que somos los organizadores de la tarea.
% [11:26, 26/4/2023] David Losada USC: porque las oraciones que se metieron en el pool vienen de las runs de los participantes
%%% Aquí se describe la colección (general stats, sentences per sympton, relación con BDI y tarea eRisk, etc.) -> TO-DO Anxo

% Comentar que estos juicios expertos son bastnate mejor que respecto a crowdsourcing como amazon mechanical turk?

This section describes the construction of \textit{DepreSym}, a resource that 
derives from Task 1 of the eRisk 2023 Lab. This is a novel task that consists of identifying sentences that 
are indicative of the presence of clinical symptoms
from the individuals who wrote these sentences. 
We follow the BDI-II, a well-studied 
clinical questionnaire, which covers 21 symptoms of depression. It includes emotional (\textit{Pessimism or Sadness}), cognitive (\textit{Indecision}) and physical (\textit{Fatigue}) symptoms~\cite{beck1996comparison}. The sentences come from a large corpus of users' posts that were written by multiple social media users, coming from the Reddit platform. The users' posts were segmented into sentences and a TREC-style collection was created (\num{3807115} sentences from \num{3107} unique users). All extracted sentences were public and Reddit terms allow the use of its contents for research purposes. 

The eRisk participants were given the full collection of sentences and were asked to submit 21 rankings of sentences (one for each BDI-II symptom) ordered by decreasing relevance to the symptom. Each participant team could submit up to 5 variants 
(runs) and each ranking had up to 1000 sentences. 
Prior to annotation, we obtained candidate sentences
by following a top-k ($k=50$) pooling approach on the submitted runs (37 runs from 10 different teams). Table \ref{tab:examples_relevant_sentences} provides two examples of candidate sentences annotated as relevant and non-relevant for the symptom \textit{Loss of energy}. Note that all sentences are somehow on-topic but only those in the lower block were labelled as relevant. These two relevant sentences offer insights into the individuals' state related to the BDI-II symptom. This stringent notion of relevance adds complexity to the labelling process. The first block of Table~\ref{table:stats_BDI_symptoms} reports the total number of annotated and relevant sentences. Here, the number of relevant sentences corresponds to the ones unanimously agreed upon by all human assessors. We can see that the number of relevant sentences is substantially low, with the $11\%$ of the sentences annotated as relevant from the pool of candidates (pool sizes ranging from \num{829} to \num{1150} sentences). The number of relevant sentences ranges from $21$ to $260$. The rest of the blocks correspond to the annotations agreement among the symptoms, explained in Subsection~\ref{subsec:symptom-based}.

\begin{table}
\setlength{\tabcolsep}{2pt}
\centering
\footnotesize
\caption{Number of sentences and annotations agreement (in percentage) statistics per symptom.}
\begin{subtable}[t]{\linewidth}
\centering
\begin{tabular}{lrrrrrrrrrrr}
    & \rot[65]{\textbf{Sadness}}
    & \rot[65]{\textbf{Pessimism}}
    & \rot[65]{\textbf{Sense of Failure}}
    & \rot[65]{\textbf{Loss of Pleasure}}
    & \rot[65]{\textbf{Guiltiness}}
    & \rot[65]{\textbf{Self-Punishment}}
    & \rot[65]{\textbf{Self-dislike}}
    & \rot[65]{\textbf{Self-incrimination}}
    & \rot[65]{\textbf{Suicidal Ideas}}
    & \rot[65]{\textbf{Crying}} 
    & \rot[65]{\textbf{Agitation}} \\
    
    \toprule
    \textbf{\# Sentences} & 1110 & 1150 & 973 & 1013 & 829 & 1079 & 1005 & 1072 & 953 & 983 & 1080 \\
    \textbf{\# Rel. sent.} & 179 & 104 & 160 & 97 & 83 & 21 & 158 & 76 & 260 & 230 & 69 \\
    
    \midrule
    \textbf{GPT-4 vs Cons.} & 72.6 & 75.7 & 73.6 & 74.6 & 75.6 & 88.2 & 72.0 & 75.6 & 77.2 & 79.8 & 80.7  \\
    \textbf{GPT-4 vs Maj.} & 76.1 & 77.1 & 76.0 & 82.0 & 81.2 & 89.3 & 81.6 & 81.3 & 85.9 & 87.7 & 80.7  \\

    \midrule

    \textbf{PhD s. vs Rest} & 80.0 & 70.5 & 84.2 & 88.5 & 90.8 & 97.0 & 81.4 & 86.5 & 87.4 & 89.3 & 88.9 \\
    \textbf{GPT-4 vs Rest} & 73.0 & 75.6 & 75.3 & 79.6 & 78.9 & 89.6 & 73.3 & 76.6 & 77.4 & 82.8 & 81.4\\   

    \midrule

    \textbf{Psy. vs Rest} & 83.0 & 76.8 & 74.1 & 82.7 & 90.0 & 95.8 & 85.3 & 86.7 & 89.8 & 88.1 & 88.4  \\
    \textbf{GPT-4 vs Rest} & 73.6 & 75.6 & 73.9 & 74.5 & 76.2 & 88.2 & 74.2 & 77.4 & 81.7 & 82.0 & 80.0   \\

    \midrule

    \textbf{Postdoc vs Rest} & 85.4 & 78.3 & 84.5 & 82.1 & 86.9 & 92.4 & 84.2 & 86.2 & 89.4 & 88.9 & 87.6 \\

    \textbf{GPT-4 vs Rest} & 74.8 & 77.3 & 74.0 & 77.2 & 77.3 & 88.4 & 78.1 & 78.5 & 81.2 & 82.5 & 80.7 \\    
    \bottomrule
\end{tabular} 
\end{subtable}
\begin{subtable}[b]{\linewidth}
\centering
\footnotesize
\begin{tabular}{lrrrrrrrrrrr}
    & \rot[65]{\textbf{Social issues}}
    & \rot[65]{\textbf{Indecision}}
    & \rot[65]{\textbf{Worthlesness}}
    & \rot[65]{\textbf{Low energy}}
    & \rot[65]{\textbf{Sleep issues}}
    & \rot[65]{\textbf{Irritability}}
    & \rot[65]{\textbf{Appetite issues}}
    & \rot[65]{\textbf{Concentration}}
    & \rot[65]{\textbf{Fatigue}} 
    & \rot[65]{\textbf{Low libido}}
    & 	\textbf{Avg.}
    
    \\

    \toprule

    \textbf{\# Sentences} & 1077 & 1110 & 1067 & 1082 & 938 & 1047 & 984 & 1024 & 1033 & 971 & \multirow{2}*{-}\\
    \textbf{\# Rel. sent.} & 70 & 61 & 71 & 129 & 203 & 94 & 103 & 83 & 123 & 97 \\
    
    \midrule
    \textbf{GPT-4 vs Cons.} & 75.9 & 79.2 & 80.9 & 78.3 & 62.1 & 84.1 & 71.7 & 83.9 & 74.8 & 81.6 & \multirow{2}{*}{-}\\
    
    \textbf{GPT-4 vs Maj.} & 81.2 & 83.3 & 84.7 & 84.5 & 76.4 & 88.0 & 80.5 & 88.2 & 83.7 & 85.6 \\

    \midrule

    \textbf{PhD s. vs Rest} & 87.7 & 90.8 & 90.3 & 88.1 & 80.7 & 86.1 & 84.6 & 90.9 & 88.2 & 92.4 & 86.9 \\
    \textbf{GPT-4 vs Rest} & 78.4 & 80.9 & 81.4 & 79.5 & 65.1 & 85.0 & 75.3 & 84.4 & 77.2 & 82.2 & 78.7\\

    \midrule

    \textbf{Psych. vs Rest} & 85.7 & 90.6 & 91.7 & 91.9 & 83.1 & 93.6 & 86.8 & 94.0 & 89.2 & 93.3 & 87.6 \\
    \textbf{GPT-4 vs Rest} & 76.6 & 80.5 & 82.9 & 80.8 & 70.0 & 85.6 & 75.2 & 86.8 & 80.0 & 84.8 & 79.1\\

    \midrule

    \textbf{Postdoc vs Rest} & 86.4 & 88.3 & 89.6 & 90.5 & 79.2 & 92.5 & 80.9 & 91.7 & 86.5 & 86.0 & 86.5 \\
    
    \textbf{GPT-4 vs Rest} & 78.0 & 80.3 & 82.3 & 80.8 & 65.4 & 85.5 & 73.3 & 84.8 & 76.3 & 81.8 & 79.0 \\

    \bottomrule
\end{tabular}
\end{subtable}
      \label{table:stats_BDI_symptoms}
\end{table}

\section{Manual Annotation Process}
\label{sec:methodology}

%\subsection{Annotation process} \label{sec:annotation}

%In Section \ref{sec:resource}, it has been elucidated that the generation of this resource was an integral part of Task 1 in the eRisk 2023 Lab, which focused on the Search for Symptoms of Depression. 

%To accomplish this task, the participants were required to provide rankings of estimated relevant sentences for each symptom of depression from the BDI Questionnaire. The organizers of the task provided the participants with a TREC formatted sentence-tagged dataset (based on past eRisk data). 

A sentence should be considered relevant only if it provides "information about the individual state related to the BDI-II symptom". To that end, we designed a set of instructions that guide the assessment process~\footnote{\url{https://erisk.irlab.org/guidelines_erisk23_task1.html}}. The guidelines were given to the human annotators and, additionally, these textual instructions were used to prompt the LLMs in our evaluation of automatic judgements. 

We selected three human assessors with different backgrounds: a field expert
(background in Psychology), a PhD student and a Postdoc (both with backgrounds in Computer Science). First, we asked them to label the pools of the first three BDI-II topics. At this point, judges were allowed to mark sentences as ``undecided''.
Next, we calculated pairwise, Cohen's Kappa to assess the agreement between single raters, and Krippendorff's Alpha for ordinal scales to evaluate the agreement between all raters. Kappa values ranged between 0.18 and 0.51 with a median of 0.38 for the three initial symptoms. Mean Krippendorff's $\alpha$ was 0.32. These Kappa values indicate low average agreement, and $\alpha$ falls below the desirable limit of $\alpha \ge 0.667$ for reliable annotations \cite{mchugh2012interrater,krippendorff2018content}. 
%\footnote{These agreement values could be overestimated, since at this point we allowed the annotators to mark some sentences as doubts. This was not allowed in the final assessment, and a binary label was always provided.}  

Next, we had a briefing with the three annotators to try to resolve ambiguities and to make the assessments more consistent. After this meeting, they were asked to relabel again the three initial symptoms. This time, ``undecided'' labels were not allowed. Cohen's Kappa ranged between 0.30 and 0.68 with a median of 0.55. Mean Krippendorff's $\alpha$ increased up to 0.51, but still below the recommendable limit. The agreement analysis for the overall assessments (over the $21$ BDI-II symptoms) led to Cohen's Kappa between $0.58$ and $0.65$ 
(median of $0.58$) and Krippendorff's $\alpha$ of $0.60$. These figures are much higher than those obtained before the briefing. This suggests that the annotation process was solid but, still, the agreement scores are moderate, reflecting the difficulty of the task. In any case, we produced two types of relevance assessments, consensus and majority, and consider the first as a high-quality container of sentences that are unambiguously relevant. We compare both assessments in the next section.

%the utility of our guidelines and the briefing process, they also demonstrate that we are facing a difficult task and it is hard to reach a high agreement between annotators. 

% \subsection{Experimental Settings}

% \begin{itemize}
%     \item Usamos un BERT  y T5 como baselines clasificacion? Pero claro, que dataset usamos? Solo el de SIGIR de Anxo + PsySym?
% \end{itemize}

%\section{LLMs for Complex Relevance Assessment}

\section{LLMs as Automatic Annotators}
\label{sec:llms-automatic-annotators}

% Agreement VINUCA: 51% and 57%
% Agreement CHATGPT: 57% and 65%
% Agreement GPT4: 77% and 83%

\begin{table}[tb]
\footnotesize
\caption{Agreement between each LLM and two types of ground truth (consensus and majority).}
\label{tbl:results-overview}
\begingroup
\begin{tabular}{@{}llcccccc}
\toprule
\textbf{LLM} & \textbf{Prediction} & \multicolumn{3}{c@{}}{\textbf{Consensus}} & \multicolumn{3}{c@{}}{\textbf{Majority}} \\
\cmidrule(l@{\tabcolsep}){3-5} \cmidrule(l@{\tabcolsep}){6-8}
&              & Rel. & Not Rel. & $\kappa$ & Rel. & Not Rel. & $\kappa$ \\
\midrule
\multirow{2}{*}{ChatGPT}
& Rel.     & \textbf{2358}      & 9277 & \multirow{2}{*}{0.18} & \textbf{4241} & 7394 & \multirow{2}{*}{0.38} \\
& Not Rel. &    113   &  \textbf{9832} & & 290 & \textbf{9655} &  \\
\midrule
\multirow{2}{*}{GPT-4}
& Rel.  & \textbf{2296} & 4755 & \multirow{2}{*}{0.38} & \textbf{3916}  & 3135 & \multirow{2}{*}{0.57} \\
& Not Rel. & 175   &  \textbf{14354} & & 615 & \textbf{13914} \\
\bottomrule
\end{tabular}
\endgroup
\end{table}

% \begin{table}[t]
%     \caption{Annotators vs The Rest.}

%     \centering
%     \begin{tabular}{@{}lcccc@{}}
%     \toprule
%      Annotator & Rest-Consensus     \\ \midrule

%      PhD student & 0.87 \\
%      ChatGPT   &  0.59 \\
%      GPT-4 & 0.79 \\
%      \midrule
%      Psy & 0.87 \\
%      ChatGPT   &  0.59 \\
%      GPT-4 & 0.79 \\
%      \midrule
%      Postdoc & 0.86 \\
%      ChatGPT & 0.60 \\
%      GPT-4 & 0.79 \\u
%     \end{tabular}
%     \label{tab:symptom-detection-results}
% \end{table}

% \begin{table}[t]
%     \caption{Ranking tau correlations of all symptoms from the human annotators.}

%     \centering
%     \begin{tabular}{@{}lcccc@{}}
%     \toprule
%      Annotators & $\tau$ \\ \midrule

%      PhD Student vs Psychologist & 0.56 \\
%      PhD Student vs Postdoc & 0.46 \\
%      Psychologist vs Postdoc & 0.69 \\

%     \bottomrule
%     \end{tabular}
% \end{table}

%%%%%%%%%%%%%%%%%%%%%%%%%%%%%%%%%%%%%

We present the results of the agreement between the human annotators and the two LLMs regarding sentences' relevance to depressive symptoms in Table~\ref{tbl:results-overview}. The analysis is conducted for the two classes
of ground truth annotations: ($i$) \textit{Consensus}, where relevant sentences are identified by all human assessors, and ($ii$) \textit{Majority}, where relevant sentences are identified by at least two human annotators.

%. Consensus was reached when all three annotators agreed on the relevant label, while majority-2 considered a sentence relevant if at least two of the three annotators labelled it as such. 
%Overall, the LLMs demonstrated significantly better performance in identifying relevant sentences compared to non-relevant ones. 
Under the consensus ground truth, ChatGPT accurately identified 95\% of the relevant sentences, compared to 93\% for GPT-4. However, both models struggled to correctly identify sentences marked as non-relevant
(accuracies of 51\% and 75\% for ChatGPT and GPT-4, respectively). This trend persists for the majority ground truth, but the correlation with human judgements shows a significant improvement. The Cohen's  
$\kappa$ level of agreement increases from $0.18$ to $0.38$ for ChatGPT and from $0.38$ to $0.57$ for GPT-4. These figures indicate a considerable increase in the performance of GPT-4 compared to ChatGPT. Another finding is that the ``non-relevant'' predictions of the models tend to be trustworthy. For instance, ChatGPT identified correctly \num{9832} out of \num{9945} non-relevant sentences. However, the predictions of relevance are much noisier. This suggests that LLMs could be the basis of a hybrid annotation approach that we will further discuss in Section \ref{sec:discussion}.

%Another interesting finding is that both models clearly mislabel a significant number of relevant sentences as non-relevant, while they hardly error when they state a sentence is not relevant. This can be the basis for some hybrids annotation approaches that we will further discuss in Section \ref{sec:discussion}.

%%% PONER ANTES EXPLICACIONES DEL RESTO DE RESULTADOS (CONSENSO GLOBAL, POR SÍNTOMA, MATRIZ DE CONFUSIÓN Y TAU)

\subsection{Symptom-based Agreement}
\label{subsec:symptom-based}
% chatgpt - 56 y 65

Table \ref{table:stats_BDI_symptoms} also displays the agreement statistics for each symptom. In the second block, we report the percentage of agreement between GPT-4 and the two classes of ground truth. The agreement percentages are generally stable across all symptoms. These percentages are higher for majority ($82.63$\% mean overall) compared to consensus ($77.04$\%). Moreover, ChatGPT achieved substantially lower agreement values ($65$\% and $56$\%, respectively)\footnote{We did not include ChatGPT's results in Table ~\ref{table:stats_BDI_symptoms} due to page limitations.}.

We further conducted a comparison between the annotations provided by GPT-4 and each human annotator, reported in the last three blocks of Table \ref{table:stats_BDI_symptoms}. 
We made pairwise comparisons between GPT-4 and each human annotator. To that end, the reference ground truth
was obtained from the consensus of the two remaining human annotators. For example, to compare the PhD student vs GPT-4 we ran them against the ground truth of relevant sentences obtained from the postdoc and the psychologist.
In all cases, each human annotator achieved a higher percentage of agreement than GPT-4. Only in one symptom (\textit{Pessimism}) against the PhD student, GPT-4 achieved a higher percentage of agreement (75.57\% vs 70.52\%, respectively). The humans led to mean scores (last column of the lower table) that were substantially higher than those achieved by GPT-4 (greater than 85\% while GPT-4 was always lower than 80\%). 
By a narrow margin, the psychologist was the human who produced superior agreement scores.

%Finally, we also computed the  rank correlation between each human assessor and GPT-4 (last column of the lower Table). The Kendall $\tau$ values indicate a \hl{strong correlation}~\cite{} between the human judgments and GPT-4, with all values being higher than $0.65$. We highlight that the psychologist achieved the highest correlation coefficient, which is promising given that is the human with higher domain expertise.

% \todo{qué sentido tiene aquí usar el rank correlation? qué rankings se están comparando? sería ordenar los BDI-II items de mayor a menor? qué sentido práctico tiene eso? no sería mejor poner simplemente la media de agreement de todas las filas?}

% The trend of ChatGPT annotations followed the same results but just a bit worse overall.

\subsection{Inter-rater Agreement and Correlation of Systems Rankings}

\subsubsection{Inter-rater Agreement} The inter-rater agreement, measured using Cohen's $\kappa$ between ChatGPT and the human annotators, ranged from 0.29 to 0.32, with a median value of 0.31. For GPT-4, the Cohen's $\kappa$ scores ranged from 0.52 to 0.54, with a median of 0.53. Additionally, Krippendorff's $\alpha$ for the combination of the three human annotators and ChatGPT was 0.40, while  Krippendorff's $\alpha$ 
with GPT-4
was 0.56. These results confirm the previous findings that GPT-4 is a more reliable annotator compared with ChatGPT.

\subsubsection{Systems Rankings Correlation} We compared the ranking of the $37$ participating search systems obtained with the official assessments (consensus of the three human assessors) against a hypothetical ranking based on assessments from a single annotator. To that end, we ranked the systems by decreasing Mean Average
Precision (MAP) and compared the rankings with
Kendall's $\tau$ and AP Correlation ($\tau_{ap}$\footnote{$\tau_{ap}$ assigns greater weight to errors made to the systems positioned higher in the ranking.})~\cite{YilmazAR08_tau_ap}.  
This analysis allows us to explore to what extent the use of a single annotator alters the system's rankings.

Looking at the results in Table~\ref{tab:ranking_correlations}, we can 
observe that GPT-4 yields a high correlation ($0.86$ and $0.81$), although lower than the
correlation levels achieved by the human 
annotators. Note that the human assessors were involved in the construction of the official qrels,
while GPT-4 was not part of the official evaluation process. The results also suggest that the assessment effort could have been reduced by involving a single human assessor. Notably, the psychologist exhibits a nearly perfect correlation with the official consensus-based ranking ($0.98$). The correlations suggest a relative order among human annotators, Psych $>$ Postdoc $>$ PhD student, which is a natural consequence of their domain knowledge and level of experience. Lastly, AP correlation and Kendall's $\tau$ show
similar trends and, thus, the rankings from individual judges do not seem to induce major swaps at the top-ranked positions.

% Correlacion de sistemas bajo los distintos qrels, si cambia mucho el tau o no. ChatGPT o GPT-4 generan qrels disintos, pero generan ranking de sistemas distintos en los 40 que tenemos?

% ver en que medida seria sustitutivo el esfuerzo humano

% ver como el disagreement afecta el ranking de sistemas

% la manera de contarlo entonces seria ver el ranking de solo un asesor vs el esfuerzo total de hacer el ranking con el input de todos los asesores, en ese sentido seria como simular el esfuerzo parcial (solo un asesor) vs hacerlo todo.

% en los datos vemos que la psicologa es la mejor etiquetadora, conocimiento mas especifico que el dominio, el pre-doc el peor.

% es facil de explicar, el consenso es simplemente los qrels con esfuerzo total (involucrar ca 3 humanos), comparas el ranking de sistemas derivado de ese esfuerzo total, con el ranking que seria de los qrels ocn un solo humano o  un solo sistema y ya esta.

% y ahi el mensaje esta claro, desi valdria ella sola practicamente, los otros no aportan practicamente valor añadido, lo cual es logico teniendo en cuenta la experiencia de cada uno

% KENDALL TAU
% cons - post 0.9459459
% cons-pre 0.93050
% cons-psy 0.98198
% cons-gpt 0.86

% TAU AP
% cons-post 0.9342302675065945
% cons-pre 0.9114552169380701
% cons-psy 0.9843235102651551
% cons-gpt 0.8279811764848588

\begin{table}[t]
        \footnotesize
        \centering
        \caption{Correlations between the official ranking of systems (consensus qrels) and
        the ranking of systems obtained from the qrels of 
        a single annotator.}
        \begin{tabular}{cccccc}
         & \multicolumn{4}{c@{}}{\textbf{Annotators}}
        \\
        \cmidrule(l@{\tabcolsep}){2-5}      
         & GPT-4 & Psychologist & Postdoc & PhD student % & Maj. vs Consensus
        \\
        \midrule 
        Kendall $\tau$ & 0.86 & \textbf{0.98} & 0.95 & 0.94 % & -
        \\
        \midrule
        $\tau_{ap}$ & 0.81 & \textbf{0.97} & 0.91 & 0.88 % & -
        \\  \hline
        \end{tabular}
        \label{tab:ranking_correlations}
\end{table}
\section{Discussion} \label{sec:discussion}

Our results suggest that LLMs are significantly better at identifying sentences marked as relevant in the ground truth compared to non-relevant ones.
This finding deviates from the tendencies observed in prior research \cite{faggioli2023perspectives}, wherein varying patterns emerged based on the specific dataset. We believe that our results give grounds to propose a new efficient hybrid labelling strategy, where LLMs act as filters that automatically remove non-relevant sentences from the pools. As shown in Table \ref{tbl:results-overview}, the ``non-relevance'' predictions of LLMs are quite accurate and, thus, the human annotation effort could be reduced to review those sentences estimated as relevant by the LLM. Thus, GPT-4 would reduce the human workload by approximately 68\%, eliminating the need of annotating around \num{15000} sentences. Considering that the average human effort per assessor was 70 hours (\num{21580} sentences), this reduction would save around 49 hours of work per human. Furthermore, reducing the burden on human annotators could potentially lead to improved annotation quality and allow for an increase in the size of the annotation pool, allowing for more documents to be reviewed.

\section{Conclusions}

%The dataset is a result of the Task 1 from the eRisk initiative. It differs from similar previous resources in that a real pool of participating systems in the task has been used to create it. 

In this paper, we presented \textit{DepreSym}, a novel resource to foster research on
new depression screening models that rely on symptom markers at sentence level. The annotated sentences were obtained from a pooling approach that utilised multiple search systems and a thorough assessment method, which involved domain experts. We also reported here our endeavours to evaluate the capabilities of LLMs as relevant sentence annotators. 
We found that these models, particularly GPT-4, are promising but still make many
false positive errors. Related to this, we further intend to explore the capabilities of other models, such as LLaMA ~\cite{DBLP:journals/corr/abs-2302-13971}, and implement hybrid annotation approaches where the LLMs act as filters of non-relevant sentences.
%\input{sections/A- Appendix}

%
% The next two lines define the bibliography style to be used, and the bibliography file.

\bibliographystyle{ACM-Reference-Format}
\bibliography{sample-base}

% Losada:

% 1.- introducción
% 2.- related work (erisk, BDI, anotación automática con LLMs, etc.)
% 3.- corpus de oraciones para fomentar investigación en síntomas de depresión
% 4.- anotación con humanos (experta en psych + otros investigadores).  procedimiento seguido: pooling, guidelines iniciales, briefing, guidelines
% definitivos, 3 queries iniciales y luego aclarar dudas, etc. estudio de agreement entre humanos
% 5.- LLMs para anotación
% 6.- discusión
% 7.- conclusiones

\end{document}